\documentclass{article}

\usepackage{PRIMEarxiv}

\usepackage[utf8]{inputenc} 
\usepackage[T1]{fontenc}    
\usepackage{hyperref}       
\usepackage{url}            
\usepackage{booktabs}       
\usepackage{amsfonts}       
\usepackage{nicefrac}       
\usepackage{microtype}      
\usepackage{amsmath}        
\usepackage{amssymb}        
\usepackage{authblk}
\usepackage{multirow}
\usepackage{fancyhdr}       
\usepackage{graphicx}       
\graphicspath{{media/}}     

\newcommand{\M}{CausalFormer}
  
\title{\M: An Interpretable Transformer for Temporal Causal Discovery
}

\author[1]{Lingbai Kong}
\author[1,3]{Wengen Li}
\author[1]{Hanchen Yang}
\author[1]{Yichao Zhang}
\author[1,3]{Jihong Guan}
\author[2]{Shuigeng Zhou}
\affil[1]{Department of Computer Science and Technology, Tongji University, Shanghai, China\\ 
\{oc371,lwengen,neoyang,yichaozhang,jhguan\}@tongji.edu.cn}
\affil[2]{School of Computer Science, Fudan University, Shanghai, China\\
sgzhou@fudan.edu.cn}
\affil[3]{Corresponding author}

\begin{document}
\maketitle

\begin{abstract}
Temporal causal discovery is a crucial task aimed at uncovering the causal relations within time series data. 
The latest temporal causal discovery methods usually train deep learning models on prediction tasks to uncover the causality between time series. They capture causal relations by analyzing the parameters of some components of the trained models, e.g., attention weights and convolution weights. However, this is an incomplete mapping process from the model parameters to the causality and fails to investigate the other components, e.g., fully connected layers and activation functions, that are also significant for causal discovery. To facilitate the utilization of the whole deep learning models in temporal causal discovery, we proposed an interpretable transformer-based causal discovery model termed \M, which consists of the causality-aware transformer and the decomposition-based causality detector. The causality-aware transformer learns the causal representation of time series data using a prediction task with the designed multi-kernel causal convolution which aggregates each input time series along the temporal dimension under the temporal priority constraint. Then, the decomposition-based causality detector interprets the global structure of the trained causality-aware transformer with the proposed regression relevance propagation to identify potential causal relations and finally construct the causal graph. Experiments on synthetic, simulated, and real datasets demonstrate the state-of-the-art performance of \M~on discovering temporal causality. Our code is available at \url{https://github.com/lingbai-kong/CausalFormer}.
\end{abstract}


\section{Introduction}
The increasing amounts of time series data initiate many studies to solve various practical issues, e.g., identifying the urban function areas~\cite{urban_function}, predicting traffic flows~\cite{STFGNN}, and forecasting weather conditions~\cite{weatherforecast}.
However, there is a desire to go beyond the direct application of time series analysis, e.g., classification and prediction, and further explore the underlying causality that drives the data variations, which could greatly benefit the detection of performance anomalies in databases~\cite{perfce}, the diagnosis of the causes of network system faults~\cite{logs}, the discovery of the factors that result in traffic congestions \cite{congestionmining}, etc. 

\begin{figure}
\centering
\includegraphics[width=0.8\linewidth]{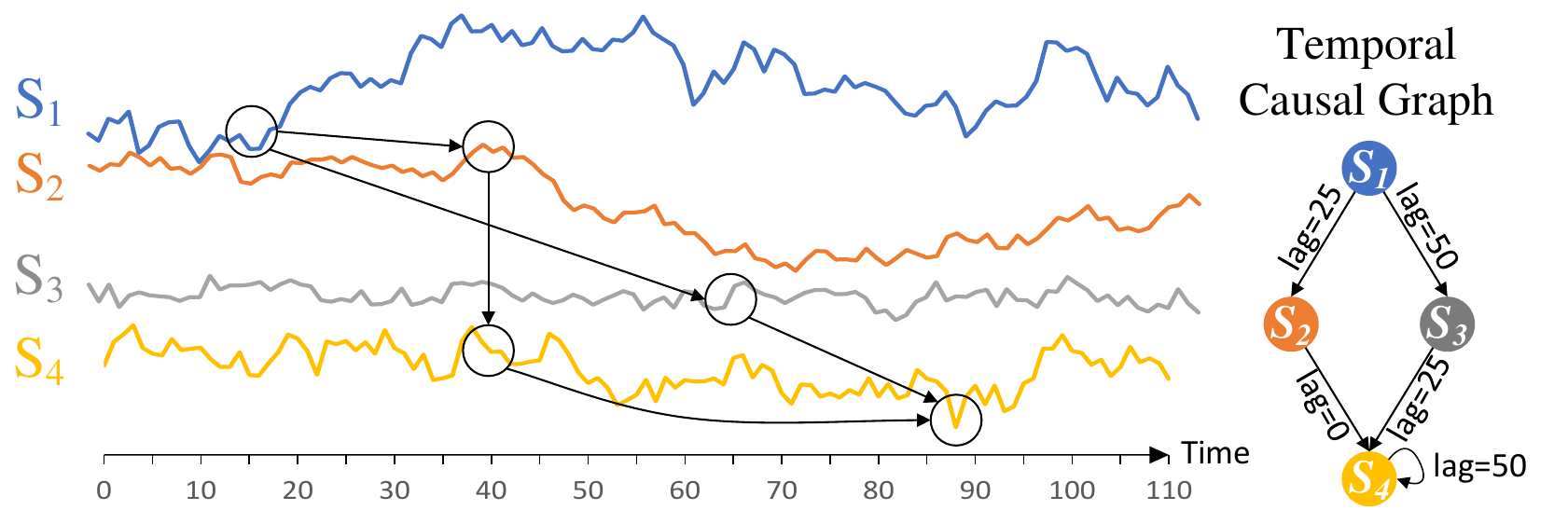}
\caption{One example of temporal causality with a diamond causal structure, where the numbers associated with edges are time lags of causal relations.}
\label{example}
\end{figure}

As a challenging yet critical task for time series data analysis, temporal causal discovery aims to identify the causality in time series data. 
Causality is referred to as the cause and effect where the cause is partly responsible for the effect while the effect is partly dependent on the cause \cite{survey_causal_inference}. 
As illustrated in Fig.~\ref{example}, there are four time series with causal relationships, where the previous values of certain time series could potentially affect the future values of other time series, and temporal causal discovery methods could construct temporal causal graphs to indicate the temporal causal relations with time lags, e.g., $S_1$$\rightarrow$$S_2$, $S_1$$\rightarrow$$S_3$ and $S_3$$\rightarrow$$S_4$. In addition, there could be instantaneous causality ($S_2$$\rightarrow$$S_4$) and self-causation ($S_4$$\rightarrow$$S_4$) in the temporal causal graph.
The learned causal graphs are beneficial for revealing the mechanism of data variation and guiding the design of time series data analysis methods.

Traditionally, researchers conduct randomized controlled trials to infer causality \cite{survey_causal_inference}, which randomly assigns participants into treatment groups and control groups to analyze the effects of various treatments and interventions in real-world settings. However, randomized controlled trials are usually infeasible due to the prohibitive cost, ethical concerns, and impracticality~\cite{survey_DAGs}. Therefore, various causal discovery approaches have been developed to solve this problem with observational data. 
Existing methods for temporal causal discovery can be roughly divided into \textit{statistic-based methods} and \textit{deep learning-based methods}.
Statistic-based methods unveil causal relations by analyzing the distribution or testing the dependence of time series data.
Deep learning-based methods train deep learning models on specific tasks and attempt to mine the causal relations by interpreting the trained models.
In practice, it is difficult for deep learning models to explicate the temporal causal relations due to their black-box characteristics~\cite{survey_interpretability}. Therefore, existing methods only map the local parameters, e.g., attention weights and convolution weights, of some components in the model to the causality for identifying causal relations. For example, Nauta et al.~proposed the Temporal Causal Discovery Framework (TCDF) \cite{TCDF} to identify the temporal causality by analyzing the temporal convolution kernel and attention weights of the trained attention-based convolutional neural networks. However, this mapping process is straightforward but incomplete~\cite{interpretable_transformer} since the rest model components, e.g., fully connected layers and activation functions, are also significant for causal representation. For instance, the multi-head attention block aggregates the input data and rudimentarily captures the latent causality. Subsequently, the fully connected layer further processes the aggregated data, where the weights concatenate the output of attention heads to refine the causal representation of the attention block, and the bias introduces the input-independent quantity to represent the latent causality, i.e., the causal relations that are not observed in input data~\cite{BBP}. Therefore, interpreting the whole structures of deep learning models is necessary for catering to their black-box characteristics and achieving complete temporal causal discovery. 

\begin{figure}
\centering
\includegraphics[width=0.8\linewidth]{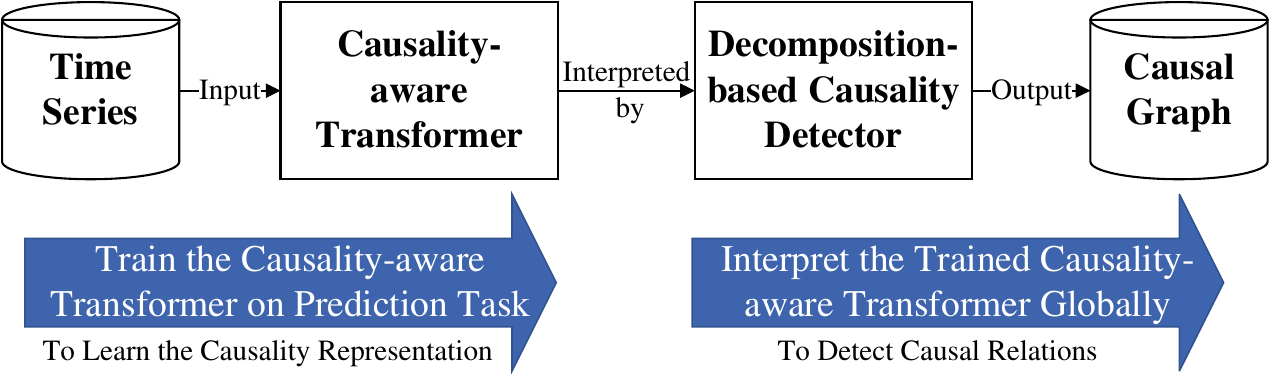}
\caption{The workflow of \M.}
\label{workflow}
\end{figure}

To address the above issue, we proposed an interpretable transformer-based model termed \M~towards temporal causal discovery. \M~consists of the causality-aware transformer and the decomposition-based causality detector. As shown in Fig.~\ref{workflow}, the causality-aware transformer is trained using a prediction task to learn the causal representation of time series. Specifically, we introduced the multi-kernel causal convolution into causality-aware transformer to aggregate each input time series along the temporal dimension under the temporal priority constraint, which helps the causality-aware transformer represent potential causal relations meticulously. Thereafter, we designed the decomposition-based causality detector to enable the global interpretability of the causality-aware transformer with the proposed regression relevance propagation that extends the layer-wised relevance propagation technique for classification models to regression models. The causality detector decomposes and backward-propagates the causal scores layer by layer, which faithfully interprets the causality-aware transformer and achieves comprehensive detection of potential causal relations. Finally, \M~constructs the causal graph of all the given time series based on the causal scores.

In sum, the major contributions of this work include:

\begin{itemize}
    \item We proposed \M, a novel deep learning-based temporal causal discovery model, consisting of the causality-aware transformer and the decomposition-based causality detector, that could learn more complete temporal causality (including self-causation and instantaneous causality) of time series by interpreting the whole structure of the trained deep learning model. 
    \item The proposed interpretable causality-aware transformer learns the representation of temporal causality with the multi-variate causal attention and multi-kernel causal convolution, and the decomposition-based causality detector achieves decomposition-based interpretation for regression models with the proposed regression relevance propagation, which enables the complete interpretability of the causality-aware transformer to calculate the causal scores of each potential causal relation and construct the temporal causal graph.
    \item Extensive experiments have been conducted on synthetic, simulated, and real datasets. The results demonstrate the state-of-the-art performance of \M~on discovering the causal relations of time series.
\end{itemize}

\section{Related Work}
In this section, we provide more details about the two categories of temporal causal discovery methods, i.e., statistic-based methods and deep learning-based methods. In addition, we also review the interpretability techniques for deep learning models since they are often used for deep learning-based causal analysis.

\subsection{Causal Discovery for Time Series}
Causal discovery for time series refers to the process of inferring causal relations from observational time series~\cite{survey_causalTS}. In comparison with the general causal discovery, the temporal causal discovery comes with the additional temporal priority constraint, i.e., the cause should occur before its effect. In some cases, the difference in time between two events associated with two time series may not be observed if the sampling frequencies are small, and the cause and effect are thus possibly observed in the same sampling time slot, i.e., instantaneous causality. Additionally, self-loop, also known as self-causation, is permitted when a time series influences its future values. 

Existing studies on causal discovery for time series can be roughly divided into statistic-based methods and deep learning-based methods.

In \textbf{statistic-based methods}, there are Granger causality methods,  constraint-based methods, noise-based methods, and score-based methods to discover temporal causal relations.
Granger causality is often used to detect causal relations by evaluating if a time series provides information for predicting the future values of other time series \cite{MVGC1,MVGC2,MVGC3}. 
It is typically studied with the linear vector autoregressive \cite{VAR} which assumes the time series at time $t$ as the linear combination of past observations, i.e., $\boldsymbol{x}^t=\sum\nolimits_{\tau =1}^T{\left\{ w_{i,j}^{\tau} \right\} \boldsymbol{x}^{t-\tau}}+\boldsymbol{e}$, where $\boldsymbol{x}^t$ is the prediction at time $t$, $\boldsymbol{x}^{t-\tau}$ is the observation at time $t-\tau$, $\left\{ w_{i,j}^{\tau} \right\}$ is the estimated matrix to aggregate the observation at time $t-\tau$, $w_{i,j}^{\tau}$ represents the causal impact of time series $i$ on time series $j$ after $\tau$ time slots, $T$ is the observation time window, and $\boldsymbol{e}$ is the residual.
Granger causality asserts that time series $i$ is the cause of time series $j$ with the delay of $\tau$ time slots if time series $i$ contributes to the prediction of time series $j$ after $\tau$ time slots \cite{granger_review}, i.e., $w_{i,j}^{\tau} \ne 0$. More extensions of Granger causality have been also proposed to solve the non-stationary and non-linear problems on real datasets \cite{nonlinearGranger,NPMR,copula}.
The constraint-based methods find causal relations between variables by exploiting their conditional independence \cite{ANLTSM,tsFCI}. The PC algorithm \cite{PC} identifies the collider (a variable is causally influenced by two variables) via independence testing, and derives other causal relations based on the inferred colliders. The PCMCI algorithm \cite{PCMCI} extends the PC algorithm to the temporal domain by conducting momentary conditional independence tests.
The noise-based methods model a causal system by a set of equations, where the effect is determined by its direct cause and additional noise \cite{noise_CI,LiNGAM}. Peters et al. proposed Time Series Models with Independent Noise (TiMINo) \cite{TiMINo} to discover causal relationships by inspecting the independence between the noise and the potential causes.  
The score-based methods model the causal graph as a probabilistic network and search for the best-match network with one certain estimation method, e.g., cross-validation estimation~\cite{CV}. Pamfil et al. designed a non-combinatorial optimization method called DYNOTEARS \cite{dynotears}, which uses adjacency matrices to represent the importance of the pairwise relation at different time lags.

\textbf{Deep learning-based methods} develop deep neural networks to represent the causality of time series and uncover the causal relations with Granger causality by quantifying the contributions of individual input observations to the model's outputs. Due to the limited interpretability of typical deep methods, it is challenging to measure the input contributions to model's prediction. Thus, researchers usually select the parameters of some components in the model to measure the input contributions. For example, cMLP~\cite{NGC} represents the input contributions with the activated neural connections of the multi-layer perceptron. 
Some models~\cite{TCDF, CGTST} focus on the information diffusion process between time series and quantify the input contributions with diffusion weights, e.g., attention weights and gate vectors. However, such contribution measurement is incomplete due to the neglect of the structure of the whole model. Therefore, more advanced interpretability techniques for deep learning models are needed to measure the input contributions accurately and further improve the performance of temporal causal discovery.

\subsection{Interpretability of Deep Learning Models}
Interpretability of deep learning models refers to the extent of human’s ability to understand the behaviors and working mechanisms of deep learning models~\cite{survey_interpretability}. Causal interpretation has been introduced in recent studies on interpretability~\cite{Gem,Causal_Screening}. 

Existing interpretability methods for deep learning models can be generally categorized into four groups as follows~\cite{survey_nn_interpretability}.
The \textbf{rule-based methods} surrogate the deep learning model with a simple and interpretable model, e.g., decision trees, to extract interpretable decision rules from the surrogate model \cite{TreeRegularization}.
The \textbf{hidden semantics-based methods} interpret the deep learning model by interpreting certain hidden neurons/layers to identify important features of input data~\cite{deepScene}.
The \textbf{example-based methods} construct the prototype inputs that make the model most accurate, e.g., constructing a prototype image that maximizes the output classification probability of a certain class \cite{prototypes}.
Regarding temporal causal discovery, however, the aforementioned interpretability methods are not applicable. The rule-based methods encounter difficulties in generating causal rules for time series. The hidden semantics-based methods do not take into account the correlations between inputs and outputs. The example-based methods are unable to provide quantitative interpretation for further building causal graphs. 

The \textbf{attribution-based methods} assign importance scores to the inputs via intervention or decomposition, e.g., masking certain inputs to test their contributions to the model's outputs, which is compatible with the Granger causality. Some methods employ gradients to indicate the importance of different input features~\cite{Grad-cam,axiomatic}. The perturbation methods monitor the changes of outputs for different input perturbations to obtain the importance scores of inputs~\cite{gnnexplainer,missingExplanations}. Additionally, the decomposition methods decompose the outputs of the deep learning model, e.g., classification probabilities, to the inputs layer by layer and regard such decomposed values as the interpretation of the model's prediction \cite{deepTaylor,DEGREE}.  
In comparison with other attribution-based methods, the decomposition approaches explain the contribution of each input to model's outputs quantificationally by analyzing most of the parameters and components of the deep learning models.

\subsection{Discussion}
After the above review, the deficiency of existing deep learning-based temporal discovery methods is attributed to that they only consider local parameters of some components in the model and neglect the structure of the whole model, which is insufficient to achieve complete temporal causal discovery. The decomposition-based interpretability techniques for deep learning models cater to the black-box characteristics of deep learning models and can be used to promote the performance of temporal causal discovery by sufficiently interpreting the global structure of the deep learning models. Therefore, in this work, we design a new decomposition-based interpretation method to release the representation power of deep learning models for discovering temporal causality. 

\begin{figure*}[t]
\centering
\includegraphics[width=\textwidth]{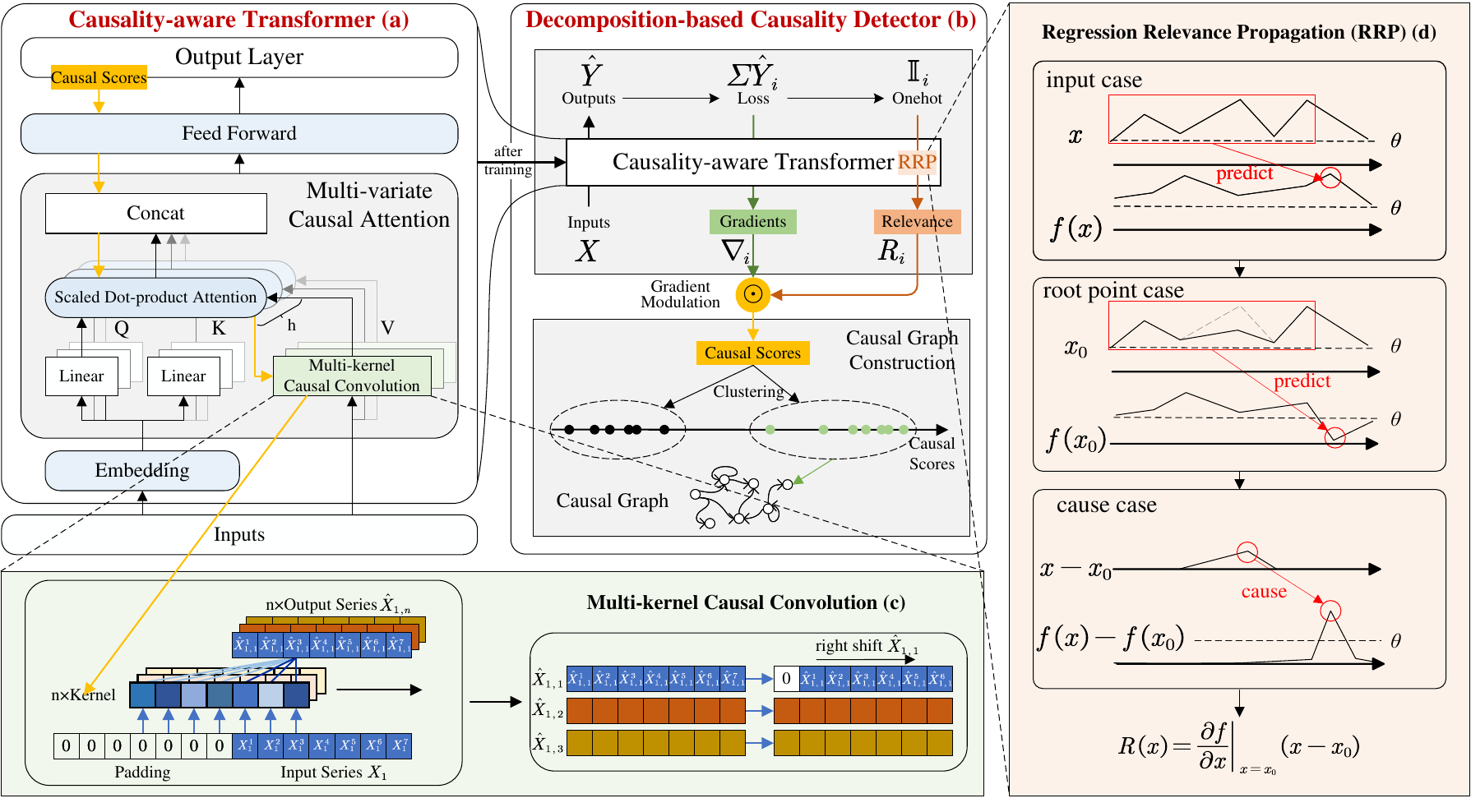}
\caption{The structure of \M, where the causality-aware transformer (a) learns the causal representation of time series with sequential layers; the decomposition-based causality detector (b) backward propagates both the relevance scores and gradients to the attention matrix and the causal convolution kernels, selects the true causal relations by clustering, and finally outputs the temporal causal graph; (c) illustrates the internal structure of multi-kernel causal convolution block; and (d) demonstrates the regression relevance propagation process for time series.
}
\label{framework}
\end{figure*}

\section{Problem Formulation}
To discover temporal causal relations among time series with the deep learning model, a prediction task on time series data is first conducted to help the deep learning model learn the representation of causality. Then, the causal discovery process leverages interpretability techniques to identify the potential causal relations from the trained deep learning model and construct the causal graph.

Concretely, given $N$ time series in an observational window of $T$ time slots, i.e., $\boldsymbol{X}$$=$$\left[ \boldsymbol{X}_1,\boldsymbol{X}_2,\cdots ,\boldsymbol{X}_N \right]$$\in$$\mathbb{R} ^{N\times T}$, we first aim to learn a function $f$ for predicting time series at the current time slot $t$ based on the previous observations as follows,
\begin{equation}
    \boldsymbol{X}_{[\![1,N]\!]\backslash \left\{ i \right\}}^{1:t},\boldsymbol{X}_{i}^{1:t-1}\xrightarrow{f}\tilde{\boldsymbol{X}}_{i}^{t},~~1\leqslant t\leqslant T
\end{equation}
where $\tilde{\boldsymbol{X}}_{i}^{t}$ is the predicted value of time series $i$ at current time slot $t$, $\boldsymbol{X}_{[\![1,N]\!]\backslash \left\{ i \right\}}^{1:t}$ represents the observations of other time series in the previous and current time slots, and $\boldsymbol{X}_{i}^{1:t-1}$ represents the previous observations of time series $i$.

Then, we wish to discover the causal relations between $N$ time series and denote them as a causal graph, i.e., $\mathcal{G}$$=$$\left( V,E \right)$, where $V$ is the set of vertices corresponding to the $N$ time series, and $E$ is the set of directed edges corresponding to the causal relations between time series. Each edge $e_{i,j}$$\in$$E$ is annotated with a weight $d(e_{i,j})$ which means the time series $i$ affects the time series $j$ with the delay of $d(e_{i,j})$ time slots.

\section{Methodology}
Fig.~\ref{framework} illustrates the overall framework of the proposed \M~which consists of the causality-aware transformer and the decomposition-based causality detector. Causality-aware transformer uses the multi-kernel causal convolution to learn the causal representation of time series under the constraint of temporal priority.
Then, the decomposition-based causality detector explores the parameters of the trained causality-aware transformer and calculates the causal scores for each potential causal relation via the regression relevance propagation.

\subsection{Causality-aware Transformer}
The causality-aware transformer is an interpretable transformer-based deep learning model trained on time series data under the constraints of temporal priority. It consists of the time series embedding layer, the multi-kernel causal convolution blocks, the multi-variate causal attention block, the feed forward layer, and the output layer. 
Firstly, the embedding layer projects each input time series into a high dimensional vector, while the multi-kernel causal convolution block convolves the previous observations of each time series with temporal priority. Secondly, the multi-variate causal attention block aggregates the convolution results with multiple attention heads according to the embeddings. Finally, the feed forward layer and the output layer further process the attention results to make predictions.

\subsubsection{Time Series Embedding}
The time series embedding layer projects the time series data into a high feature space. Specifically, given the time series $\boldsymbol{X}$$\in$$\mathbb{R} ^{N\times T}$, the time series embedding layer projects the row vector $\boldsymbol{X}_i$$\in$$\mathbb{R} ^{1\times T}$ to the dimension $d$ and $d>T$, i.e.,
\begin{equation}\boldsymbol{X}_{emb}=\boldsymbol{X}\times \boldsymbol{W}_{emb}+\boldsymbol{b}_{emb}\end{equation}
where $\boldsymbol{W}_{emb}\in \mathbb{R} ^{T\times d}$ and $\boldsymbol{b}_{emb}$ are the weights and biases, and $\boldsymbol{X}_{emb}\in \mathbb{R} ^{N\times d}$ is the embedding of $N$ input time series. The embedding is only used by query $\boldsymbol{Q}$ and key $\boldsymbol{K}$ in the multi-variate causal attention block because the embedding blends the time series across the time dimension, and the value $\boldsymbol{V}$ must keep the temporal order of the input time series to satisfy the temporal priority constraint.

\subsubsection{Multi-kernel Causal Convolution}
The multi-kernel causal convolution block aggregates the historical and current observations of input time series for prediction with multiple kernels under the temporal priority constraint to establish independent series-to-series data flow and help the causality-aware transformer represent potential causal relations meticulously. Specifically, a causal convolution kernel is defined as a learnable matrix. i.e., $\boldsymbol{\mathcal{K}}$$\in$$\mathbb{R} ^{N\times N\times T}$, where the three dimensions correspond to the time series to be convolved, the time series to be predicted, and the convolution field of $T$ time slots, respectively.
Moreover, we left pad the input time series with a zero vector to fill the values of previous time slots. Take $\boldsymbol{X}_1$ as an example, as shown in Fig.\ref{framework} (c), when conducting convolution on $\boldsymbol{X}_1$ at time $t$, we first left pad the input data $\boldsymbol{X}_1$$=$$\left[ \boldsymbol{X}_{1}^{1},\cdots ,\boldsymbol{X}_{1}^{T} \right]$ with a $T$-length zero vector, i.e., $\left[ \boldsymbol{0}^1,\cdots ,\boldsymbol{0}^T,\boldsymbol{X}_{1}^{1},\cdots ,\boldsymbol{X}_{1}^{T} \right]$. Then we convolute the padded input vector with the corresponding kernel $\boldsymbol{\mathcal{K}}_1$ to aggregate the historical information for prediction at time slot $t$, i.e., 
\begin{equation}\hat{\boldsymbol{X}}_{1}^{t}=\boldsymbol{\mathcal{K}} _1\cdot \left[ \boldsymbol{0}^{t+1},\boldsymbol{0}^T,\boldsymbol{X}_{1}^{1},\cdots ,\boldsymbol{X}_{1}^{t} \right] /t\end{equation}
where $\hat{\boldsymbol{X}}_{1}^{t}$$\in$$\mathbb{R} ^{1 \times N \times1}$ is the convolution result, and $\left[ \boldsymbol{0}^{t+1},\boldsymbol{0}^T,\boldsymbol{X}_{1}^{1},\cdots ,\boldsymbol{X}_{1}^{t} \right]$ is the sub-vector of the padded vector that has the same size of the convolution kernel and contains only historical information to satisfy the temporal priority constraint. The number of non-zero elements in the sub-vector, i.e., $t$, is used to scale the convolution result. The complete convolution result is denoted as $\hat{\boldsymbol{X}}$$=$$\left[ \hat{\boldsymbol{X}}_1,\cdots ,\hat{\boldsymbol{X}}_N \right] \in \mathbb{R} ^{N\times N\times T}$, and the vector $\hat{\boldsymbol{X}}_{i,j}\in \mathbb{R} ^{1 \times 1 \times T}$ denotes the convolution result of time series $i$ for predicting time series $j$ and represents the temporal causal relation from time series $i$ to $j$. 

Additionally, the multi-kernel causal convolution also identifies the instantaneous self-causation by right shifting the self-convolution result $\hat{\boldsymbol{X}}_{i,i}$, i.e.,
\begin{equation}\left[ \hat{\boldsymbol{X}}_{i,i}^{1},\cdots ,\hat{\boldsymbol{X}}_{i,i}^{T} \right] \xrightarrow{shift}\left[ \boldsymbol{0},\hat{\boldsymbol{X}}_{i,i}^{1},\cdots ,\hat{\boldsymbol{X}}_{i,i}^{T-1} \right]\end{equation}
The groundtruth is thus not exposed in the prediction, which ensures the correctness of self-causation learning.

\subsubsection{Multi-variate Causal Attention}
The multi-variate causal attention block learns the complex causal correlations between time series. We first define the query $\boldsymbol{Q}$, key $\boldsymbol{K}$, and value $\boldsymbol{V}$ as follows,
\begin{equation}\boldsymbol{Q}=\boldsymbol{X}_{emb}\times \boldsymbol{W}_Q+\boldsymbol{b}_Q,\boldsymbol{K}=\boldsymbol{X}_{emb}\times \boldsymbol{W}_K+\boldsymbol{b}_K,\boldsymbol{V}=\hat{\boldsymbol{X}}\end{equation}
where $\boldsymbol{W}_Q,\boldsymbol{W}_K\in \mathbb{R} ^{d\times d_{QK}}$ and $\boldsymbol{b}_Q,\boldsymbol{b}_K$ are the weights and biases to project the embedding to dimension $d_{QK}$, and $\boldsymbol{V}$ directly employs the results of causal convolution to keep the temporal priority constraint. Then, the multi-variate causal attention is written as
\begin{equation}\boldsymbol{A}=\underset{\boldsymbol{\mathcal{A}}}{\underbrace{\mathrm{soft}\max \left( \frac{\boldsymbol{Q}\times \boldsymbol{K}^T}{\tau \cdot \sqrt{d_{QK}}}\odot \boldsymbol{M} \right) }}\times \boldsymbol{V}\end{equation}
where $\boldsymbol{M}$$\in$$\mathbb{R} ^{N\times N}$ is the learnable attention mask to adjust the sparsity of the causal graph by $L_1$ normalization, and $\tau$ is the temperature hyperparameter that modulates the probability distribution of softmax. The output of softmax is referred to as the attention matrix $\boldsymbol{\mathcal{A}} \in \mathbb{R} ^{N\times N}$. The attention result $\boldsymbol{A}\in \mathbb{R} ^{N\times T}$ is calculated by $\boldsymbol{A}_{i,t}=\sum\nolimits_{j=1}^N{\boldsymbol{\mathcal{A}} _{ij}\cdot \boldsymbol{V}_{j,i,t}}$. Then, we duplicate the multi-variate causal attention $h$ times to create $h$ attention heads which are aggregated as follows.
\begin{equation}\boldsymbol{Att}=\left[ \boldsymbol{A}^{\left( 1 \right)},\cdots \boldsymbol{A}^{\left( h \right)} \right] \times \boldsymbol{W}_O\end{equation}
where $\boldsymbol{A}^{\left( \cdot \right)}$$\in$$\mathbb{R} ^{N\times T}$ denotes the attention result of a single attention head,  $\boldsymbol{W}_O$$\in$$\mathbb{R} ^{h}$ is the weight for concatenating the outputs of $h$ attention heads, and $\boldsymbol{Att}\in \mathbb{R} ^{N\times T}$ is the final output of the multi-variate causal attention. 

\subsubsection{Feed Forward Layer}
The feed forward layer is composed of two linear neural networks, separated by a leaky ReLU activation function  in the middle, structured as follows,
\begin{equation}FFN\left( \boldsymbol{Att} \right) =\mathrm{Linear}\left( \mathrm{leakyReLU}\left( \mathrm{Linear}\left( \boldsymbol{Att} \right) \right) \right) \end{equation}

Feed forward layer transforms the layer input dimension to $d_{FFN}$ and then restores it to the original dimension, thus introducing non-linearity to enhance the causal representability for the causality-aware transformer.

\subsubsection{Output Layer}
The output layer is a fully connected layer following the feed forward layer to make the prediction $\tilde{\boldsymbol{X}}$$\in$$\mathbb{R} ^{N\times T}$. Mean squared error (MSE) is used to optimize the learnable parameters of the model. Practically, we ignore the prediction of the first time slot for the sake of fairness because the observations of each time series do not contribute to their own predictions in the first time slot due to the right shifting of self-convolution result. Furthermore, the loss function $\mathcal{L}$ encourages sparsity of the causality with the $L1$ normalization on the causal convolution kernels and the attention mask, i.e.,
\begin{equation}\begin{split}\mathcal{L} \left( \varTheta \right) &=\frac{\sum{\left( \tilde{\boldsymbol{X}}^{[\![1,T]\!]\backslash \left\{ 1 \right\}}-\boldsymbol{X}^{[\![1,T]\!]\backslash \left\{ 1 \right\}} \right) ^2}}{N\cdot T}\\
&+\lambda_{\boldsymbol{\mathcal{K}}} \cdot \left\| \boldsymbol{\mathcal{K}} \right\| _1+\lambda_{\boldsymbol{M}} \cdot \left\| \boldsymbol{M} \right\| _1\end{split}\end{equation}
where $\varTheta$ denotes all learnable parameters in the causality-aware transformer, $\tilde{\boldsymbol{X}}^{[\![1,T]\!]\backslash \left\{ 1 \right\}}$ is the predictions except the first time slot, $\boldsymbol{X}^{[\![1,T]\!]\backslash \left\{ 1 \right\}}$ is the observations/groundtruth except the first time slot, $\left\| \cdot \right\| _1$ denotes the $L1$ normalization, $\lambda_{\boldsymbol{\mathcal{K}}}$ and $\lambda_{\boldsymbol{M}}$ are the normalization coefficients, $\boldsymbol{\mathcal{K}}$ is the causal convolution kernel, and $\boldsymbol{M}$ is the attention mask.

\begin{figure*}[!t]
\centering
\includegraphics[width=0.9\textwidth]{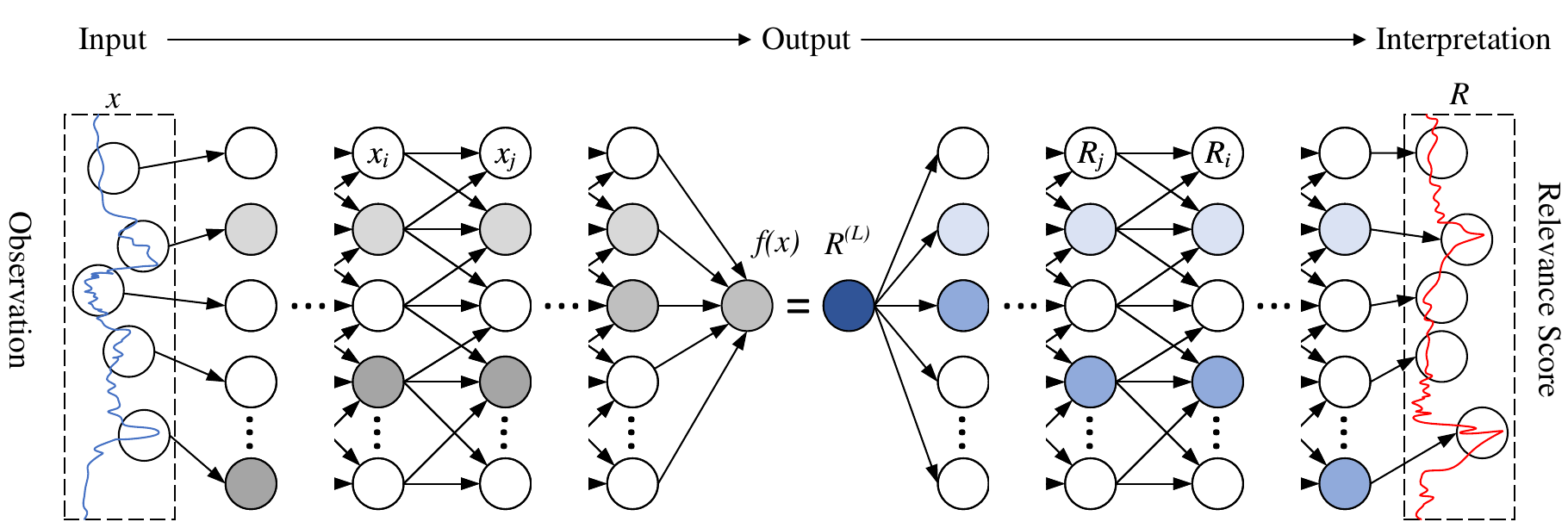}
\caption{The computational flow of layer-wised relevance propagation (LPR) on single time series input. The prediction function $f(\boldsymbol{x})$ first makes the classification with the given observation $\boldsymbol{x}$. Then the output neuron is assigned relevance $\boldsymbol{R}^{(L)}$$=$$f(\boldsymbol{x})$, which is decomposed as a sum of terms called relevance scores $\boldsymbol{R}$ and backward propagated to the neurons of each layer. The neurons with a high relevance score are colored in blue.}
\label{LRP}
\end{figure*}

\subsection{Decomposition-based Causality Detector}
The causality-aware transformer learns the causal representation of time series data. Then, the decomposition-based causality detector interprets the causality-aware transformer globally to discover the temporal causal relations by backward decomposing the output predictions to the attention matrix and the causal convolution kernels with the proposed regression relevance propagation (RRP) to detect causal relations and calculate the corresponding causal delays.

\subsubsection{Regression Relevance Propagation} As shown in Fig. \ref{LRP}, layer-wised relevance propagation (LPR) \cite{LPR} defines a decomposition-based interpretation method for deep classification models. It decomposes the prediction function $f\left( \boldsymbol{x} \right)$ as a sum of terms called relevance scores of the separate input nodes of each layer, i.e., 
\begin{equation}f\left( \boldsymbol{x} \right) =\sum_{d\in V_{L}}{\boldsymbol{R}_{d}^{\left( L \right)}}=\cdots =\sum_{d\in V_l}{\boldsymbol{R}_{d}^{\left( l \right)}}=\cdots =\sum_{d\in V_1}{\boldsymbol{R}_{d}^{\left( 1 \right)}}\label{Eq:LPR}\end{equation}
where $V_l$ denotes the set of input nodes of the $l$-th layer, and $\boldsymbol{R}_{d}^{(l)}$ denotes the relevance score of the $d$-th node in the $l$-th layer. The relevance score measures the contribution of the input of a certain layer to the output prediction. According to Granger causality, a larger positive relevance score indicates a causal relation between the input time series and the predicted time series. 

Deep Taylor decomposition \cite{deepTaylor} implements the LPR with the first-order Taylor approximation which is applicable for general differentiable classifiers. For the single-layer neural network of $V$ input nodes, the first-order Taylor expansion of the prediction function is given as,
\begin{equation}f\left( \boldsymbol{x} \right) =f\left( \boldsymbol{x}_0 \right) +\sum_{d=1}^V{\underset{\boldsymbol{R}_d}{\underbrace{\left. \frac{\partial f}{\partial \boldsymbol{x}_{\left( d \right)}} \right|_{\boldsymbol{x}=\boldsymbol{x}_0}\cdot \left( \boldsymbol{x}_{\left( d \right)}-\boldsymbol{x}_{0\left( d \right)} \right) }}}+\boldsymbol{\varepsilon }\label{Eq:taylor}\end{equation}
where $\boldsymbol{x}_0$ is the root point near input $\boldsymbol{x}$ and $f(\boldsymbol{x}_0)=0$, $\boldsymbol{x}_{\left( d \right)}-\boldsymbol{x}_{0\left( d \right)}$ is the difference between the input and the root point on the $d$-th input node, the summation terms are identified as the relevance scores $\boldsymbol{R}_d$ of the $d$-th input node, and the term $\boldsymbol{\varepsilon}$ denotes the higher-order Taylor approximation terms.

For the classification model, the prediction output indicates the probability of each class that a given input belongs to. The root point $\boldsymbol{x}_0$ plays the role of a blank sample that is similar to the original input $\boldsymbol{x}$ but lacks the key information that causes $f(\boldsymbol{x})$ to be positive. For the regression model, however, the output of the model is the prediction value, and the root point does not make sense when the groundtruth is zero. Thus, the regression relevance propagation (RRP) zooms the data values larger than a positive threshold $\theta $ as shown in Fig.~\ref{framework} (d). For the input case, the regression model uses the input time series data $x$ to make predictions which is larger than the threshold $\theta $. Then the root point $\boldsymbol{x}_0$ is defined to leave the prediction drop to zero in the root point case. In this way, $f(\boldsymbol{x}_0)$$=$$0$ means the blank prediction without key information, and in the cause case, $\boldsymbol{x}-\boldsymbol{x}_0$ denotes the causal part of the input time series data to the prediction results.

\begin{figure}[t]
\centering
\includegraphics[width=0.4\linewidth]{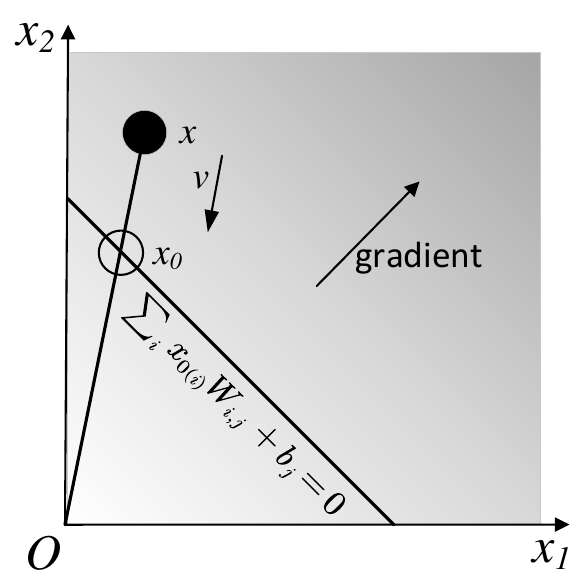}
\caption{The searching process of the nearest root point $\boldsymbol{x}_0$ (empty cycle) of the interpreted input $\boldsymbol{x}$ (solid cycle).}
\label{x0}
\end{figure}

To search the nearest root point $\boldsymbol{x}_0$ of the interpreted input $\boldsymbol{x}$, we need to find the intersection of the root plane equation and a line indicating the root point searching direction. As illustrated in Fig.~\ref{x0}, for the fully connected layer, the equations of the root plane on the $j$-th output node and the searching line can be defined as follows,
\begin{equation}
\begin{cases}
	\sum_i{\boldsymbol{x}_{0\left( i \right)}\cdot \boldsymbol{W}_{i,j}}+\boldsymbol{b}_j=0\\
	{\boldsymbol{x}_0}_{\left( i \right)}=\boldsymbol{x}_{\left( i \right)}+r\cdot \boldsymbol{v}_i,r\in \mathbb{R} \,\,i\in V\\
\end{cases}
\label{Eq:P&L}
\end{equation}
where $\boldsymbol{x}_{0\left( i \right)}$ and $\boldsymbol{x}_{\left( i \right)}$ are the elements of $\boldsymbol{x}_0$ and $\boldsymbol{x}$ corresponding to the $i$-th input node, $\boldsymbol{W}_{i,j}$ and $\boldsymbol{b}_j$ are the weight and bias, respectively, $\boldsymbol{v}_i$ represents the elements of the chosen direction vector $\boldsymbol{v}$, and $r$ is a scalar variable representing the position along the searching line. Then, we solve Eq. \eqref{Eq:P&L} for $\boldsymbol{x}-\boldsymbol{x}_0$ and rewrite the relevance score of Eq. \eqref{Eq:taylor} as,
\begin{equation}\boldsymbol{R}_{i}^{(l)}=\sum_j{\frac{\boldsymbol{v}_i\cdot \boldsymbol{W}_{i,j}}{\sum_{i\prime}{\boldsymbol{v}_{i\prime}\cdot \boldsymbol{W}_{i\prime,j}}}\cdot}\boldsymbol{R}_{j}^{(l+1)}\end{equation}
where $\boldsymbol{R}^{(l)}_i$ is the current layer relevance score of the $i$-th input node, and $\boldsymbol{R}^{(l+1)}_j$ is the upper layer relevance score of the $j$-th output node.  

There are many ways to choose the direction vector. Here we use z-rule \cite{LPR} which defines $\boldsymbol{v}_i$$=$$\boldsymbol{x}_{(i)}$, i.e., the direction vector from the origin point to $\boldsymbol{x}$. Thus, the relevance score under z-rule is defined as below.
\begin{equation}\boldsymbol{R}_{i}^{(l)}=\sum_j{\frac{\boldsymbol{x}_{\left( i \right)}\cdot \boldsymbol{W}_{i,j}}{\sum_{i\prime}{\boldsymbol{x}_{\left( i\prime \right)}\cdot \boldsymbol{W}_{i\prime,j}}}\cdot}\boldsymbol{R}_{j}^{(l+1)}\end{equation}
Here, $\boldsymbol{R}_i^{(l)}$ is the weighted average of $\boldsymbol{R}_j^{(l+1)}$ based on $\boldsymbol{x}_{\left( i \right)} \cdot \boldsymbol{W}_{i,j}$. Considering that bias is also meaningful for interpreting the causality-aware transformer, we add bias to the denominator of the weighted term and define the relevance score and bias relevance as follows.
\begin{equation}\boldsymbol{R}_{i}^{(l)}=\sum_j{\frac{\boldsymbol{x}_{\left( i \right)}\cdot \boldsymbol{W}_{i,j}}{\sum_{i\prime}{\boldsymbol{x}_{\left( i\prime \right)}\cdot \boldsymbol{W}_{i\prime,j}}+\boldsymbol{b}_j}}\cdot \boldsymbol{R}_{j}^{(l+1)}
\label{Eq:baisedRelevance}\end{equation}
\begin{equation}\boldsymbol{R}_{\left[ \boldsymbol{b}_j \right]}^{(l)}=\sum_j{\frac{\boldsymbol{b}_j}{\sum_{i\prime}{\boldsymbol{x}_{\left( i\prime \right)}\cdot \boldsymbol{W}_{i\prime,j}}+\boldsymbol{b}_j}}\cdot \boldsymbol{R}_{j}^{(l+1)}\end{equation}
The bias relevance $\boldsymbol{R}_{\left[ \boldsymbol{b}_j \right]}^{(l)}$ subtracts a portion of $\boldsymbol{R}^{(l+1)}_j$, which violates Eq. \eqref{Eq:LPR}  that requires the sum of relevance scores to be equal at each layer. However, this violation is beneficial for temporal causal discovery. A larger $\boldsymbol{R}_{\left[ \boldsymbol{b}_j \right]}^{(l)}$ means that the bias is a causal factor influencing the prediction, consequently resulting in a reduction in $\boldsymbol{R}_i^{(l)}$, and thus downsizing the corresponding causality.

Then, we replace $\sum_{i\prime}{\boldsymbol{x}_{\left( i\prime \right)}\cdot \boldsymbol{W}_{i\prime,j}}+\boldsymbol{b}_j$ with the output value for the $j$-th output node in the $l$-th layer, i.e.,  $f^{(l)}(\boldsymbol{x})_j$, and replace $\boldsymbol{W}_{i,j}$ with the partial derivative of the $j$-th output node with respect to the $i$-th input node, i.e., $\partial f^{\left( l \right)}\left( \boldsymbol{x} \right) _j/\partial \boldsymbol{x}_{\left( i \right)}$, for Eq. \eqref{Eq:baisedRelevance}. Finally, the RRP between layers is defined as follows.
\begin{equation}\boldsymbol{R}_{i}^{\left( l \right)}=\sum_j{\boldsymbol{x}_{\left( i \right)}\cdot \frac{\partial f^{\left( l \right)}\left( \boldsymbol{x} \right) _j}{\partial \boldsymbol{x}_{\left( i \right)}}\cdot \frac{\boldsymbol{R}_{j}^{\left( l+1 \right)}}{f^{\left( l \right)}\left( \boldsymbol{x} \right) _j}}\label{RRP}\end{equation}
where $\boldsymbol{R}_{i}^{\left( l \right)}$ is the relevance score of the $i$-th input node in the $l$-th layer, and $f^{(l)}$ is the layer function. This equation is applicable to any parametric layer in the causality-aware transformer. 

The other non-parametric operations, e.g., matrix production, are decomposed by non-parametric relevance propagation~\cite{interpretable_transformer} to propagate the relevance through both input tensors. Specifically, for the matrix production, assuming that the input matrices are $\boldsymbol{A}\in \mathbb{R} ^{N\times K}$ and $\boldsymbol{B}\in \mathbb{R} ^{K\times M}$ and the output matrix is $\left( \boldsymbol{A}\times \boldsymbol{B} \right) \in \mathbb{R} ^{N\times M}$, the relevance propagation is defined as below.
\begin{equation}\begin{cases}
	{R_{n,k}^{\boldsymbol{A}}}^{\left( l \right)}=\sum_m{\boldsymbol{A}_{n,k}\cdot \frac{\partial \left( \boldsymbol{A}\times \boldsymbol{B} \right) _{n,m}}{\partial \boldsymbol{A}_{n,k}}\cdot \frac{R_{n,m}^{\left( l-1 \right)}}{\sum_{k\prime}{\boldsymbol{A}_{n,k\prime}\cdot \boldsymbol{B}_{k\prime,m}}}}\\
	{R_{k,m}^{\boldsymbol{B}}}^{\left( l \right)}=\sum_n{\boldsymbol{B}_{k,m}\cdot\frac{\partial \left( \boldsymbol{A}\times \boldsymbol{B} \right) _{n,m}}{\partial \boldsymbol{B}_{k,m}}\cdot \frac{R_{n,m}^{\left( l-1 \right)}}{\sum_{k\prime}{\boldsymbol{A}_{n,k\prime}\cdot \boldsymbol{B}_{k\prime,m}}}}\\
\end{cases}\end{equation}
The equations is the variant of Eq. \eqref{RRP}, where $R_{n,m}^{\left( l-1 \right)}$ is the relevance score of  product matrix $\left( \boldsymbol{A}\times \boldsymbol{B} \right) _{n,m}$ in the ($l-1$)-th layer, and ${R_{n,k}^{\boldsymbol{A}}}^{\left( l \right)}$ and ${R_{k,m}^{\boldsymbol{B}}}^{\left( l \right)}$ are the decomposed relevance scores of operands $\boldsymbol{A}_{n,k}$ and $\boldsymbol{B}_{k,m}$ in the $l$-th layer, respectively. 

\begin{figure*}[!t]
\centering
\includegraphics[width=0.8\linewidth]{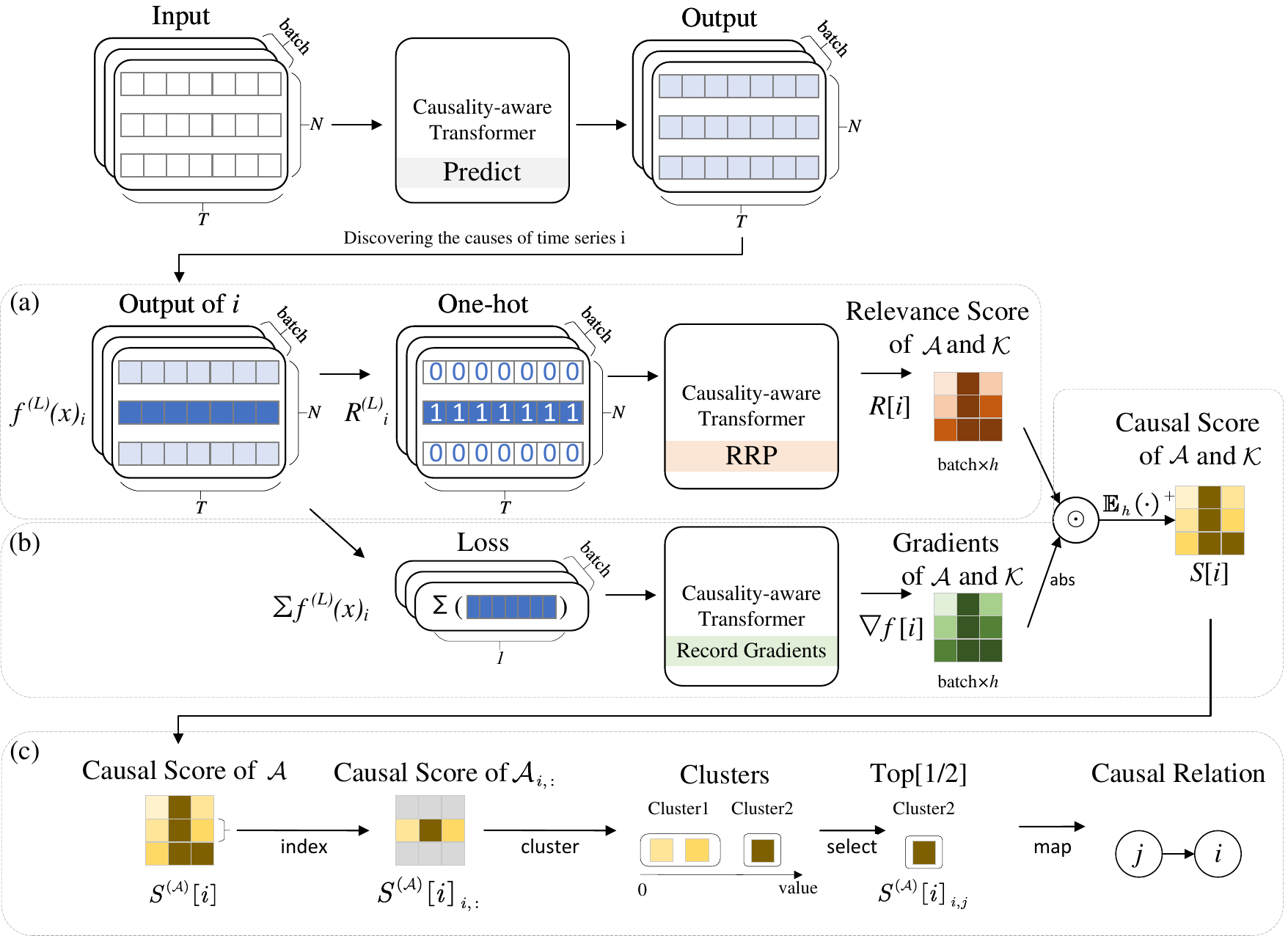}
\caption{The overall process of decomposition-based causality detector. First, the trained causality-aware transformer outputs the prediction for the given input time series. Then the regression relevance propagation decomposes the causality-aware transformer and outputs the relevance score of the attention matrix $\boldsymbol{\mathcal{A}}$ and causal convolution kernel $\boldsymbol{\mathcal{K}}$ (a). The gradient modulation strengthens the relevance scores of inputs with large absolute values of gradients to calculate the causal scores (b). Finally, the causal graph is constructed with the top classes of clustered causal scores (c).}
\label{CausalScore}
\end{figure*}

As shown in Fig. \ref{CausalScore} (a), to discover the causes of time series $i$, we initialize the relevance scores of the output layer $\boldsymbol{R}^{\left( L \right)}$ with a one-hot vector, i.e.,  $\left[ \boldsymbol{0}^1,\cdots ,\boldsymbol{0}^{i-1},\boldsymbol{1}^i,\boldsymbol{0}^{i+1},\cdots ,\boldsymbol{0}^N \right]$ to standardize the initial relevance scores of batch samples.

\subsubsection{Gradient Modulation}
Gradients are also important to indicate the causal relations between time series because the outputs are sensitive to those inputs with large absolute values of gradients. Therefore, the relevance score is modulated by gradient as illustrated in Fig. \ref{CausalScore} (b), i.e.,
\begin{equation}\boldsymbol{S}_{d}^{\left( l \right)}=\mathbb{E} _h\left( \left| \boldsymbol{\nabla} f_{d}^{\left( l \right)} \right|\odot \boldsymbol{R}_{d}^{\left( l \right)} \right) ^+\end{equation}
where $\boldsymbol{S}_d^{(l)}$, $\boldsymbol{\nabla} f_{d}^{\left( l \right)}$, and $\boldsymbol{R}_{d}^{\left( l \right)}$ are respectively the causal score, gradient, and relevance score of the $d$-th input node in the $l$-th layer, $(\cdot)^+$ is the rectification operator that sets all negative values to zero, and $\mathbb{E} _h$ is the mean operation across attention heads. Only the positive causal scores are taken into consideration, thus disregarding the non-causal relations.

\subsubsection{Causal Graph Construction}
In practice, we conduct the decomposition process from the output layer, passing through the fully connected layers, and culminating in the attention matrix $\boldsymbol{\mathcal{A}}$ and causal convolution kernel $\boldsymbol{\mathcal{K}}$ within the the multi-variate causal attention block for the convenience of mapping the causal scores of $\boldsymbol{\mathcal{A}}$ and $\boldsymbol{\mathcal{K}}$ to potential causal relations and potential causal delays, respectively. The ignored embedding layer and $QK$ projection layers have limited impacts on the causal discovery results because they do not integrate information across the time series data and thus cannot represent causal relations.

When discovering the causes of time series $i$, the causal score of the attention matrix is denoted by $\boldsymbol{S}^{\left( \boldsymbol{\mathcal{A}} \right)}\left[ i \right] \in \mathbb{R} ^{N\times N}$, and the causal score of the causal convolution kernel is denoted by $\boldsymbol{S}^{\left( \boldsymbol{\mathcal{K}} \right)}\left[ i \right] \in \mathbb{R} ^{N\times N\times T}$. As illustrated in Fig. \ref{CausalScore} (c), we first select the causal scores relevant to the time series $i$, i.e., $\boldsymbol{S}^{\left( \boldsymbol{\mathcal{A}}  \right)}\left[ i \right] _{i,:}$ and $\boldsymbol{S}^{\left( \boldsymbol{\mathcal{K}} \right)}\left[ i \right] _{:,i,:}$. Then, we cluster the elements of $\boldsymbol{S}^{\left( \boldsymbol{\mathcal{A}}  \right)}\left[ i \right] _{i,:}$ into $n$ classes with k-means \cite{kmeans}, and sort the classes according to their centroids. The causal score in the top $m$ class, i.e., $\boldsymbol{S}^{\left( \boldsymbol{\mathcal{A}}  \right)}\left[ i \right] _{i,j}\in \mathrm{Top}\left[ {{m}/{n}} \right]$, indicates the causal relation from time series $j$ to $i$, and the corresponding causal graph edge $e_{j,i}$ is created. The $m/n\in \left[ 0,1 \right] $ controls the density of the causal graph and a larger $m/n$ results in a denser causal graph. The largest kernel causal score of the cause time series $j$ indicates the lag of the causal relation, i.e. 
\begin{equation}
d\left( e_{j,i} \right) =T-\underset{t}{\mathrm{argmax}}\left( \boldsymbol{S}^{\left( \boldsymbol{\mathcal{K}} \right)}\left[ i \right] _{j,i,t} \right)
\end{equation}
Finally, \M~outputs the complete temporal causal graph of all the given time series.

\section{Experiments}
\subsection{Datasets}
We validate our model on six datasets, including four synthetic datasets, the simulated climate dataset Lorenz96, and the real fMRI neuroscience dataset.

\begin{figure}[h]
\centering
\includegraphics[width=0.6\linewidth]{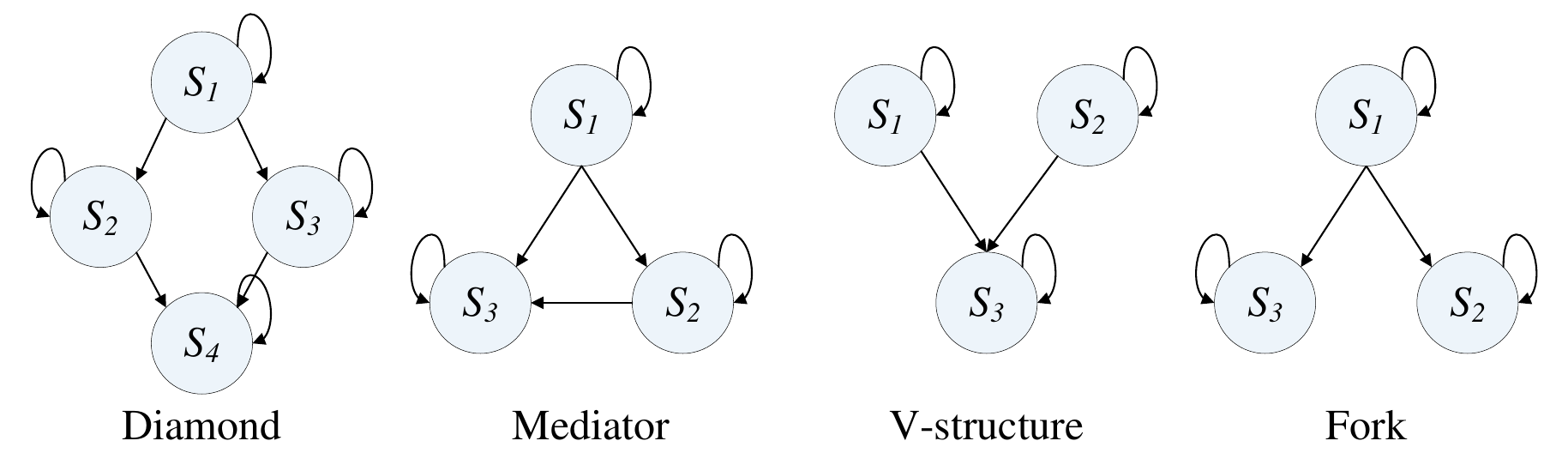}
\caption{The causal graphs of four types of synthetic datasets, i.e., diamond, mediator, v-structure, and fork.}
\label{basic}
\end{figure}

\textbf{Synthetic datasets:} As illustrated in Fig.~\ref{basic}, these datasets construct different basic causal structures, i.e., diamond, mediator, v-structure, and fork, with additive noise of a standard normal distribution. The diamond dataset contains four time series while each of the other three synthetic datasets contains three time series. The length of each time series is set to 1,000. 

\textbf{Lorenz 96 dataset:} This dataset is generated by using the Lorenz 96 model which is a nonlinear model of climate dynamics~\cite{lorenz} as defined below.
 \begin{equation}\frac{d\boldsymbol{x}_{t,i}}{dt}=(\boldsymbol{x}_{t,i+1}-\boldsymbol{x}_{t,i-2})\cdot \boldsymbol{x}_{t,i-1}-\boldsymbol{x}_{t,i}+F\end{equation}
where $\boldsymbol{x}_{t,i}$ is the data of time series $i$ at time slot $t$, and $F$ is a forcing constant that determines the level of non-linearity and the chaos in the time series. We simulate a Lorenz-96 model with 10 variables and $F\in \left[ 30,40 \right] $ over a time span of 1,000 units.

\textbf{fMRI dataset:} Functional magnetic resonance imaging (fMRI) dataset contains the BOLD (blood-oxygen-level dependent) data for 28 different brain networks \cite{fMRI}. The time series in the fMRI dataset represent the neural activities of spatial regions of interest (ROIs).
The time spans of these time series are different and fall within the range of 50 to 5,000. The numbers of monitoring time series in the 28 brain networks are also different and could be 5, 10, 15, or 50. 

In addition, to further evaluate the applicability of CausalFormer, we apply it to detect the causal relations in a sea surface temperature (SST) data which contains 260 time series of length 97. 




\subsection{Baselines}

\begin{itemize}
    \item \textbf{cMLP\&cLSTM (2021) \cite{NGC}:} cMLP and cLSTM are neural Granger causal discovery models which identify the causality of time series by training and interpreting the multi-layer perceptron (MLP) model and the long short-term memory (LSTM) model, respectively.
    \item \textbf{TCDF (2019) \cite{TCDF}:} The temporal causal discovery framework (TCDF) applies convolution neural network (CNN) and attention mechanism to learn the causal relations between time series. 
    \item \textbf{DVGNN (2023) \cite{DVGNN}:} The dynamic diffusion-variational graph neural network (DVGNN) constructs causal graphs for time series with graph convolution network (GCN) and diffusion model. 
    \item \textbf{CUTS (2023) \cite{CUTS}:} Causal discovery from irregular time series (CUTS) discovers the causality from irregular time series data by imputing unobserved data points with delayed supervision graph neural network (DSGNN), and building the causal graph under sparse penalty. 
\end{itemize}

\subsection{Experimental Settings}
We briefly introduce the training scheme, experiment environment, and evaluation metrics as follows.

\textbf{Training scheme:} 
To balance the performance and time consumption, we carefully tune hyper-parameters for each dataset as follows. 
For the synthetic datasets with fewer time series, we configure \M~with $d_{QK}$$=$$d$$=$$256$, $h$$=$$4$, $d_{FFN}$$=$$256$, $m/n$$=$$1/2$, $T$$=$$16$. We set $\tau$$=$$1$ and $\lambda_{\boldsymbol{\mathcal{K}}}$=$\lambda_{\boldsymbol{M}}$$=$$10^{-4}$ for diamond and mediator. Because the non-self causal relations are less than causal relations in v-structure and fork, we set $\tau $$= $$100$ and $\lambda_{\boldsymbol{\mathcal{K}}}$=$\lambda_{\boldsymbol{M}}$$= $$10^{-10}$ to improve the causal discovery ability for non-self causal relations. For the Lorenz dataset, we set $d_{QK}$$=$$d$$=$$512$, $h$$=$$8$, $d_{FFN}$$=$$512$, $\tau$$=$$10$, $\lambda_{\boldsymbol{\mathcal{K}}}$=$\lambda_{\boldsymbol{M}}$$=$$5\times 10^{-4}$, $m/n$$=$$2/3$, and $T$$=$$32$ to handle the dense and non-linear causal relations. For the fMRI dataset, we set $d_{QK}$$=$$d$$=$$256$, $h$$=$$4$, $d_{FFN}$$=$$512$, $\tau$$=$$100$, $m/n$$=$$1/2$, and $T$$=$$32$ to handle more complex causal relations. We remove the normalization items by setting $\lambda_{\boldsymbol{\mathcal{K}}}$=$\lambda_{\boldsymbol{M}}$$=$$0$ to encourage the discovery of more causal relationships. The model parameters are initialized by He initialization \cite{HeInit} and optimized by Adam with the early stop strategy. All the baselines run with the author-provided configuration. Since DVGNN and CUTS output the causal scores for each potential causal relation, we also identify the causal relations by k-means as \M.

\textbf{Experiment environment:} 
All the experiments run on an Ubuntu server with NVIDIA 4090 GPU 24GB and Intel(R) Xeon(R) Gold 6226R CPU@2.90GHz.

\textbf{Evaluation metrics:} 
We use the standard precision, recall, and F1-score to evaluate the performance of temporal causal discovery methods, and use the precision of delay (PoD) to evaluate the performance of causal delay discovery.

\begin{table*}[t]
\caption{The overall F1-score (mean ± standard deviation) of \M~and baseline methods. The highest score per benchmark is highlighted in bold.
\label{table_overview}}
\renewcommand\arraystretch{1.2}
\centering
\resizebox{!}{!}{\begin{tabular}{cc|cccccc}
\hline
\multicolumn{2}{c|}{Dataset}             & cMLP               & cLSTM     & TCDF               & DVGNN     & CUTS      & \M                 \\ \hline
\multirow{4}{*}{Synthetic} & Diamond     & 0.55±0.19          & 0.63±0.13 & \textbf{0.68±0.09} & 0.65±0.04 & 0.49±0.20 & \textbf{0.68±0.08} \\
                           & Mediator    & \textbf{0.71±0.14} & 0.59±0.24 & 0.69±0.06          & 0.65±0.05 & 0.52±0.23 & \textbf{0.71±0.06} \\
                           & V-structure & 0.73±0.15          & 0.60±0.20 & 0.76±0.09          & 0.73±0.06 & 0.49±0.15 & \textbf{0.77±0.05} \\
                           & Fork        & 0.51±0.33          & 0.47±0.32 & 0.73±0.10          & 0.75±0.00 & 0.50±0.19 & \textbf{0.79±0.11} \\ \cline{1-2}
Simulated                  & Lorenz96    & 0.64±0.03          & 0.63±0.06 & 0.46±0.05          & 0.48±0.07 & 0.58±0.02 & \textbf{0.69±0.06} \\ \cline{1-2}
Realistic                  & fMRI        & 0.58±0.14          & 0.56±0.13 & 0.59±0.12          & 0.56±0.12 & 0.61±0.13 & \textbf{0.66±0.09} \\ \hline
\end{tabular}}
\end{table*}
\normalsize

\subsection{Results and Discussion}
Table~\ref{table_overview} presents the overall results of \M~and multiple baseline methods on different datasets.
When comparing the F1-scores on the synthetic datasets, our proposed model exhibits superior performance to other deep learning-based methods, which indicates that \M~can accurately identify various basic causal structures. As for the experiment results on the simulated and real datasets, i.e., Lorenz96 and fMRI, \M~outperforms all the other baselines, which shows that our method can fully leverage the powerful representation and learning ability of deep learning models to detect temporal causality.
 
Causal delay discovery is another challenge for deep learning-based temporal causal discovery models. Our model can also output the time lags of discovered causal relations, which is a bonus of the multi-kernel causal convolution. As shown in Table~\ref{table_PoD}, we only compare the precision of delay (PoD) of cMLP and TCDF with \M~since the other baselines do not output the time lags of the discovered causal relations. 
Meanwhile, the results on the fMRI dataset are also not reported since it does not contain the groundtruth of causal delay. 
According to the results, the PoD of \M~is inferior, which is because cMLP imposes more penalties to more previous observations, and TCDF detects the time delay for each causal relation with the dilated temporal convolution, while our model fairly employs the observations of the whole time window. For future work, the constraint or penalty on the causal convolution process is worth exploring to improve the PoD while maintaining the performance of temporal causal discovery.

\begin{table}[t]
\caption{The PoD of cMLP, TCDF, and \M.
\label{table_PoD}}
\renewcommand\arraystretch{1.2}
\centering
\begin{tabular}{cccc}
\hline
            & cMLP      & TCDF      & Ours      \\ \hline
Diamond     & 0.82±0.17 & 0.92±0.13 & 0.74±0.20 \\
Mediator    & 0.91±0.12 & 0.97±0.11 & 0.63±0.40 \\
V-structure & 0.91±0.16 & 1.00±0.00 & 0.59±0.39 \\
Fork        & 0.76±0.41 & 1.00±0.00 & 0.46±0.34 \\
Lorenz96    & 0.45±0.17 & 0.77±0.08 & 0.42±0.18 \\ \hline
\end{tabular}
\end{table}

\begin{table}[t]
\caption{The results of different variants of \M~on the fMRI dataset, where w/o means removing the corresponding component.
\label{table_ablation}}
\renewcommand\arraystretch{1.2}
\centering
\begin{tabular}{lccc}
\hline
\textbf{Experiment}            & \textbf{Precision} & \textbf{Recall}    & \textbf{F1}        \\ 
\hline
w/o interpretation    & 0.47±0.24 & 0.45±0.17 & 0.44±0.18 \\
w/o relevance         & 0.64±0.32 & 0.44±0.12 & 0.50±0.17 \\
w/o gradient          & 0.60±0.60 & 0.54±0.54 & 0.54±0.54 \\
w/o bias              & 0.79±0.31 & 0.44±0.12 & 0.55±0.18 \\
w/o multi conv kernel & 0.74±0.25 & 0.56±0.12 & 0.61±0.12 \\
\M                    & \textbf{0.80±0.17} & \textbf{0.59±0.13} & \textbf{0.66±0.09} \\
\hline
\end{tabular}
\end{table}

\subsection{Ablation Studies}

The ablation studies are conducted on the fMRI dataset. We compare the overall precision, recall, and F1-score on different variants of \M, including: 
(1) w/o interpretation, which removes the decomposition-based causality detector and regards attention matrix and convolution kernel weights of causality-aware transformer as the causal scores; 
(2) w/o relevance, which removes relevance scores and regards the absolute values of gradients  as the causal scores; 
(3) w/o gradient, which removes gradient modulation and regards relevance scores as the causal scores; 
(4) w/o bias, which removes the bias item in RRP; 
(5) w/o multi conv kernel, where the multi-kernel causal convolution block captures information with a single convolution kernel.

The results are shown in Table \ref{table_ablation}. Our proposed RRP plays a critical role in temporal causal discovery. Concretely, removing the interpretation processing has the greatest impact on the causal discovery performance, highlighting the essential role of the proposed decomposition-based causality detector in identifying the potential causal relations from the trained deep learning model. The relevance decomposition focuses more on the recall while the gradients are more conducive to the accuracy. The bias item in RRP helps avoid incorrectly attributing the contributions of some biases to the prediction of input time series, which further improves the recall. Multiple convolution kernels help establish independent series-to-series data flow for meticulous causality representation, which contributes to the accuracy and recall.

\begin{figure}
\centering
\includegraphics[width=0.8\linewidth]{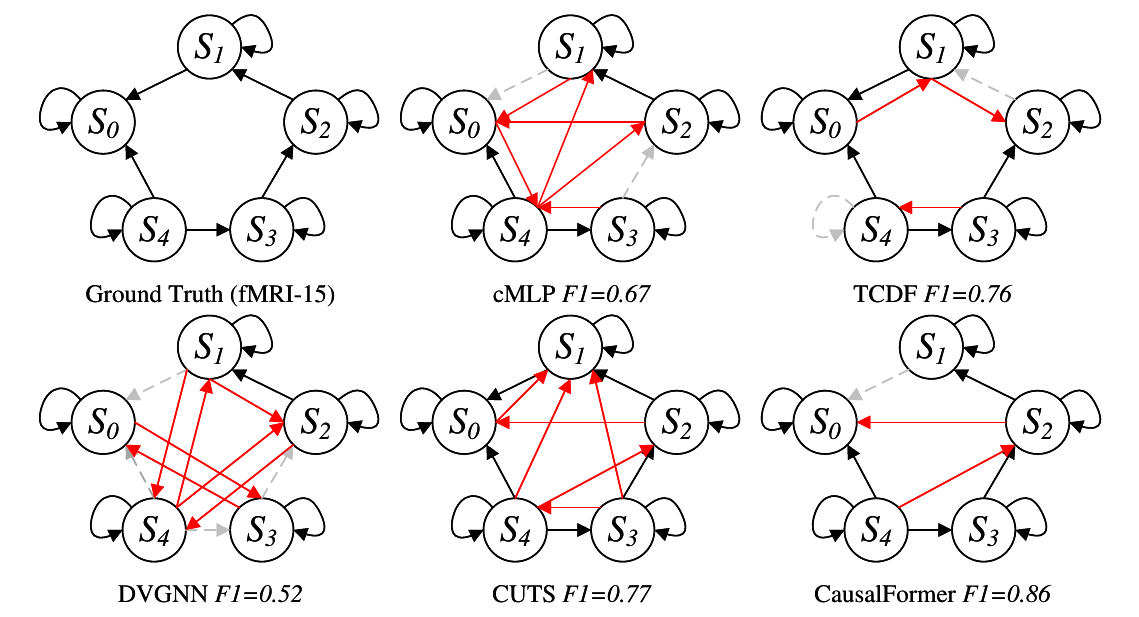}
\caption{The groundtruth causal graph and the detected causal relations by different methods on the fMRI-15 dataset, where black edges, red edges, and dashed edges denote the true positive causal relations, false positive causal relations, and false negative causal relations, respectively.}
\label{case}
\end{figure}

\subsection{Case Study}
Fig.~\ref{case} visualizes the causal graphs of the groundtruth and the causal discovery results on the fMRI-15 dataset. \M~misidentifies two indirect causal relations as causal relations and ignores one causal relation, while the other models make more mistakes and even get the opposite causal relations to the groundtruth. For example, cMLP, TCDF, and CUTS claim that time series $S_3$ causes the time series $S_4$, which is opposite to the ground truth $S_4$$\rightarrow $$S_3$. This shows the superiority of our model in temporal causal discovery.

\begin{figure}
\centering
\includegraphics[width=0.8\linewidth]{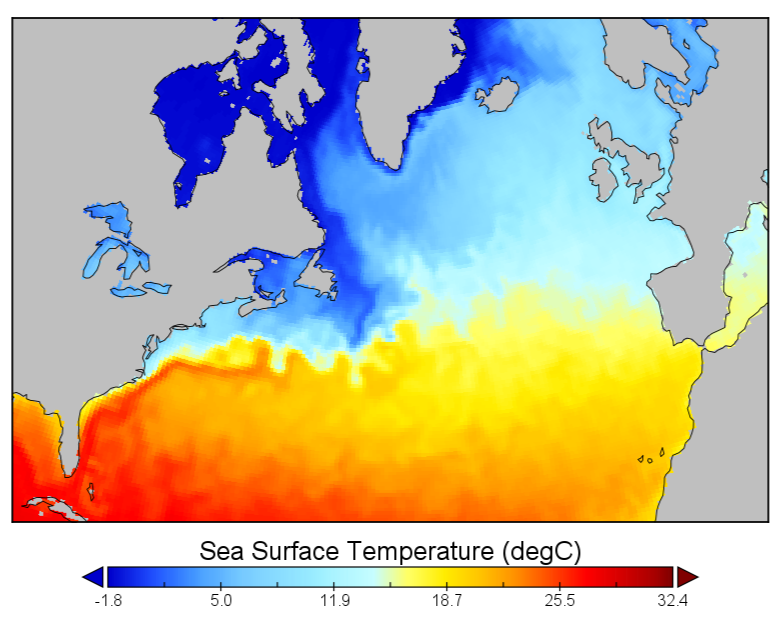}
\caption{The SST in the North Atlantic region on Jan 1st, 2022.}
\label{SSTNA}
\end{figure}

To further evaluate the applicability of CausalFormer in practice, we apply it to detect the long-term causal relations of sea surface temperature (SST) in the North Atlantic region (20°~N--70°~N,\ 0°~W--80°~W). SST is the key driving factor of climate change. Thus, exploring the causality of SST variation is significant for various meteorological applications, e.g., weather forecasting and marine ecological protection.
The SST dataset is from the OI-SST (Optimum Interpolation SST) data repository provided by NOAA (National Oceanic and Atmospheric Administration) and has a spatial resolution of 4.0°×4.0° after pre-processing. Fig.~\ref{SSTNA} shows the SST on Jan 1st, 2022 in the North Atlantic region.


\begin{figure}
\centering
\includegraphics[width=0.8\linewidth]{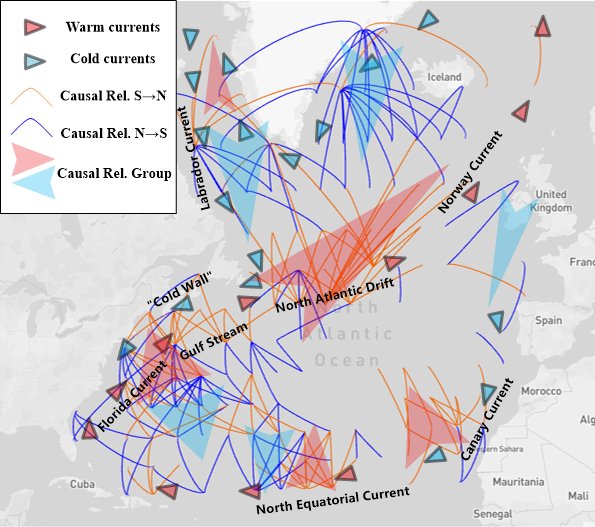}
\caption{The ocean currents and the causal discovery results on the SST data in the North Atlantic region, where the S→N (N→S) causal relations denote cause with a lower (higher) latitude than its effect.}
\label{causalitySST}
\end{figure}

We collect the SST from 2013 to 2022. According to previous studies~\cite{heattransport,CurrentEffect}, SST is greatly influenced by the ocean currents and the typical surface speed in the North Atlantic is about 0.3 $m/s$. Therefore, we choose 38 days as the span of time slot to set the movement range of ocean currents in a time slot to about 1000 kilometers and align the model with the long-term causality driven by ocean currents. As illustrated in Fig.~\ref{causalitySST}, the causal relations of SST in the North Atlantic region generally match the spatial distribution of the North Atlantic Current. Specifically, there are many S→N causal relations along the North Atlantic Drift and the Norway Current, which denotes that these warm currents carry the heat from the low latitude sea areas and cause the SST variation of high latitude sea areas. Additionally, the N→S causal relations around Greenland demonstrate the causality of the cold currents to the surface temperature of the low latitude sea areas. Moreover, it can be found that the causal relations in the western North Atlantic are much more complex than those in the eastern North Atlantic, which can be attributed to the interactions of the Gulf of Mexico and the western North Atlantic, and the sluggish flow of the Canary Current (less than 0.25m/s). In conclusion, \M~can generally identify causal relationships consistent with the ocean currents, which confirms its effectiveness and applicability.


\section{Conclusion}
This work develops a novel model termed \M~towards causal discovery on time series. Particularly, we propose the causality-aware transformer to capture the causal patterns and introduce the multi-kernel causal convolution to aggregate each time series along the temporal dimension under the temporal priority constraint. Moreover, we devise the decomposition-based interpretability technique termed regression relevance propagation, which expands the layer-wise relevance propagation method to the regression task and enables the global interpretability of the causality-aware transformer to construct the causal graph. According to the experiment evaluation on multiple datasets, \M~achieves state-of-the-art performance in discovering the causal relations of time series. The case studies on the fMRI-15 dataset and SST dataset also indicate that \M~is of high applicability.

\section*{Acknowledgments}
This work was supported in part by National Key R\&D Program of China (No. 2021YFC3300300), National Natural Science Foundation of China (No. 62202336, No. U1936205, No. 62172300, NO. 62372326), and Open Research Projects of Zhejiang Lab (No. 2021KH0AB04).

\bibliographystyle{unsrt}  
\bibliography{templateArxiv}  

\end{document}